\documentclass{article}
\usepackage{ijcai24}

\usepackage{times}
\usepackage{soul}
\usepackage{url}
\usepackage[hidelinks]{hyperref}
\usepackage[utf8]{inputenc}
\usepackage[small]{caption}
\usepackage{graphicx}
\usepackage{amsmath}
\usepackage{amsthm}
\usepackage{booktabs}
\usepackage{algorithm}
\usepackage{algorithmic}
\usepackage[switch]{lineno}
\usepackage{amssymb}
\usepackage{bbm}
\usepackage{multirow}

\title{
DDF: A Novel Dual-Domain Image Fusion Strategy for Remote Sensing Image Semantic Segmentation with Unsupervised Domain Adaptation
}

\author{
Lingyan Ran$^1$ \and
Lushuang Wang$^1$ \and
Tao Zhuo$^{*1}$ \and
Yinghui Xing$^{1}$\\
\affiliations
$^1$School of Computer Science, Northwestern Polytechnical University\\
\emails
\{lran,xyh\_7491\}@nwpu.edu.cn,
wanglushuang@mail.nwpu.edu.cn,
zhuotao724@gmail.com
}

\begin{document}

\maketitle

\begin{abstract}
Semantic segmentation of remote sensing images is a challenging and hot issue due to the large amount of unlabeled data. Unsupervised domain adaptation (UDA) has proven to be advantageous in incorporating unclassified information from the target domain.
However, independently fine-tuning UDA models on the source and target domains has a limited effect on the outcome.
This paper proposes a hybrid training strategy as well as 
a novel dual-domain image fusion strategy that effectively utilizes the original image, transformation image, and intermediate domain information. 
Moreover, to enhance the precision of pseudo-labels, we present a pseudo-label region-specific weight strategy. 
The efficacy of our approach is substantiated by extensive benchmark experiments and ablation studies conducted on the ISPRS Vaihingen and Potsdam datasets.
\end{abstract}

\section{Introduction}

\begin{figure}[t]
    \includegraphics[width=2.3in]{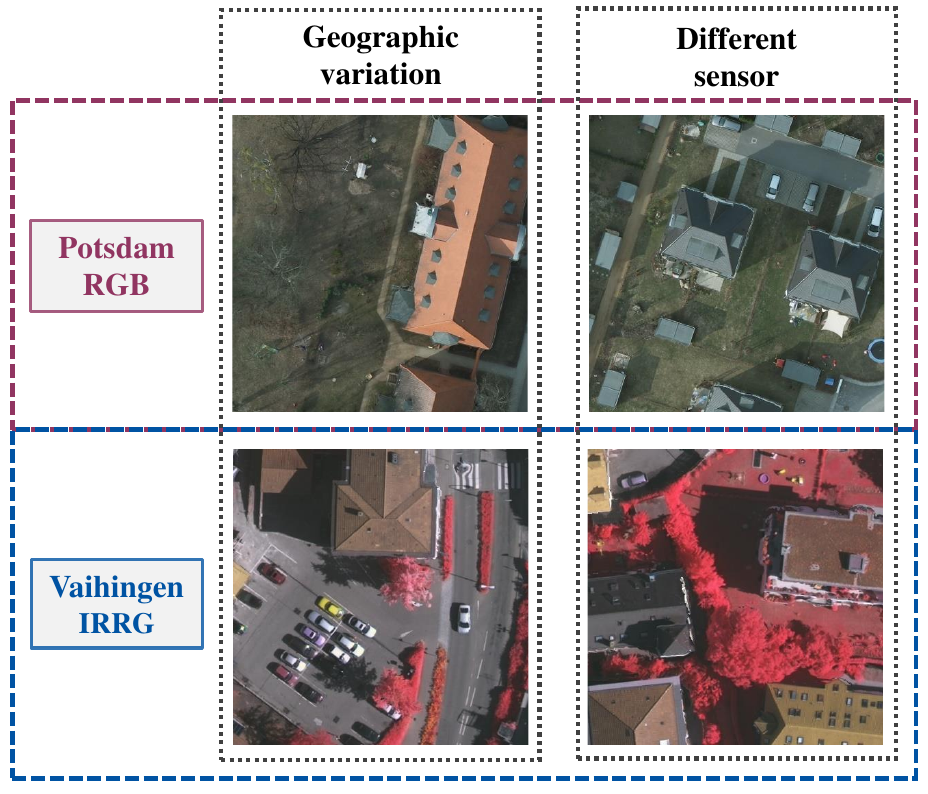}
    \centering
    \caption{The use of UDA for remote sensing images encounters numerous obstacles because of variations in geographical locations, temporal changes, and seasonal inconsistencies. Moreover, the utilization of diverse sensors leads to images that possess distinct spectral characteristics, resolutions, and levels of noise. These inherent dissimilarities among datasets contribute to the challenge of models trained on one dataset being able to generalize effectively to another.}
    \label{fig:domain}
\end{figure}

In recent times, the field of satellite and imaging technologies has made great strides, resulting in the collection of a large amount of remote sensing (RS) images. 
One crucial application in this field is semantic segmentation, such as land use analysis, road network extraction, and building inspection, which involves accurately labeling each pixel in an image. 
The process of annotating remote sensing images is both time-consuming and labor-intensive, and is further complicated by the considerable variations in the data, as illustrated in Figure~\ref{fig:domain}.
Therefore, effectively utilizing extensive unlabeled data and addressing the performance degradation caused by data discrepancies are important and difficult problems to solve.

In order to tackle these challenges, researchers have proposed the use of unsupervised domain adaptation (UDA). 
This approach involves utilizing labeled data from a source domain to train a network and make predictions for unlabeled data from a target domain. 
However, directly applying this approach to target domain data is difficult due to significant differences between datasets. 
Challenges in UDA for RS images occur because of differences in geographic locations, temporal variations, and seasonal fluctuations. These factors lead to disparities in domains across regions, causing differences in lighting conditions, types of features, soil coverage, and other environmental factors. Furthermore, the use of different sensors produces images with diverse spectral properties, resolutions, and levels of noise. 
The difficulty lies in the insufficient adaptation of models trained on one dataset to another dataset due to their inherent dissimilarities.
As a result, several methods have been introduced to mitigate the domain-gap problem. 
The primary methods include generative training (GT) methods (CycleGAN~\cite{cyclegan}, ColorMapGAN~\cite{colormapgan}), adversarial methods (AdaptSeg~\cite{adaptseg}, ADVENT~\cite{advent}), and self-training methods (CBST~\cite{cbst}, DAFormer~\cite{daformer}).

The primary focus of generative training is to modify the appearance of an image in order to reduce visual differences, such as color and texture, between images from the source and target domains. This approach aims to address the problem of domain shift at the input level. Adversarial training methods, on the other hand, are mainly used for feature-level and pixel-level adversarial training. Feature-level domain adaptation involves measuring the distance between features and reducing it, while pixel-level domain adaptation focuses on aligning the output to minimize the difference between prediction results of source and target domains. The self-training method is another approach that involves generating pseudo-labels for unlabeled target images, which are then used to guide the model for fine-tuning. Each of these mainstream approaches has its own challenges. The generative training method primarily enhances the visual aspect of the image, while the self-training method has shown better performance than adversarial methods but raises issues related to obtaining efficient pseudo-labels and effectively utilizing them for self-training on the target domain.


This paper proposes a hybrid training method to address the challenges mentioned above. The method combines the self-training approach with supplementary generative training. The generative training method is used to modify the image style while preserving the original semantic information. The self-training method is enhanced with a dual-domain image fusion module (DDF) to improve the precision of pseudo-labels. The DDF merges two image styles to create a new image that incorporates both styles, thereby reducing the domain gap. 
Additionally, the paper suggests using pseudo-label regional weights (PRW) to better utilize these labels. This involves evaluating the quality of pseudo-labels and assigning weights to different regions based on the difficulty of the categories.

To summarize, the main contributions are as follows:
\begin{itemize}
    \item We utilize a hybrid training strategy that focuses on a self-training framework, supplemented by a generative training method. By combining these two approaches strategically, we are able to reduce the negative effects of noise that may arise from the GT method. As a result, the accuracy of the pseudo-labels is enhanced, enabling more efficient utilization of these labels. 
    
    \item We propose a novel DDF module for image fusion that addresses the domain gap issue. Our module combines images from both the source and target domains, creating intermediate domain information. We perform alignment at the input level and reduce the impact of noise by filtering the generated image. Furthermore, the PRW module provides additional support for pseudo-labels.
    Ablation experiments demonstrate a significant improvement in results.
    
    \item Our method surpasses existing techniques by achieving a mIoU of 65.71\% and an F1-score of 80.10\% when carrying out the segmentation task from Potsdam R-G-B to Vaihingen. These findings indicate a substantial enhancement of 4.3\% and 5.18\% compared to the current state-of-the-art method, emphasizing the efficacy of our approach.
\end{itemize}

\section{Related Work}
\subsection{Semantic Segmentation}
In recent years, there has been significant progress in the field of semantic segmentation. This progress can be attributed to the introduction of advanced architectures, innovative techniques, and improved performance levels. ~\cite{long2015Fully} proposed a groundbreaking fully convolutional neural network (FCN) architecture, which differed from the traditional approach of using fully connected layers. This marked a turning point in the field. Subsequently, convolutional neural networks (CNNs) have become the dominant paradigm. ~\cite{dilate} introduced dilated convolutions as a means to efficiently capture contextual information at multiple scales. 
The UNet~\cite{ronneberger2015Unet} architecture and its various variants, such as UNet++~\cite{zhou2018unet++}, have been developed to enhance the segmentation capabilities of the original architecture. Most networks now follow an encoder-decoder architecture, which enables the capture of hierarchical features and the refinement of segmentation maps.

Although CNNs have been successful in extracting local features, they have limitations in capturing long-range dependencies and global contextual information, which are crucial for visual tasks. To address this, researchers have looked to natural language processing (NLP) for inspiration and adapted the Transformer architecture~\cite{transformer} for computer vision tasks like semantic segmentation. The self-attention mechanism in Transformers allows the model to assign different weights to different positions in the input sequence, effectively capturing both local and global dependencies. As a result, there is a growing trend in computer vision to explore hybrid architectures that combine the strengths of both CNNs and Transformers.

\subsection{Unsupervised Domain Adaptation}
Unsupervised domain adaptation deals with the problem of adjusting a model that has been trained on a source domain, where labeled data is accessible, to perform effectively on a target domain, where labeled data is not accessible. In UDA, the source and target domains typically exhibit dissimilar distributions, leading to a domain shift. UDA techniques can be categorized into generative training methods, adversarial-based methods, and self-training methods.

The objective of adversarial training methods is to align the distributions of the source and target domains~\cite{adv1,adv2}. One example is the domain-invariant representation, which involves a game of least-maximum adversarial optimization~\cite{CLAN}. In this game, feature extractors try to deceive domain discriminators in order to achieve aligned feature distributions. However, these adversarial training methods often have inconsistent performance.

\begin{figure*}[t]
    \includegraphics[width=5in]{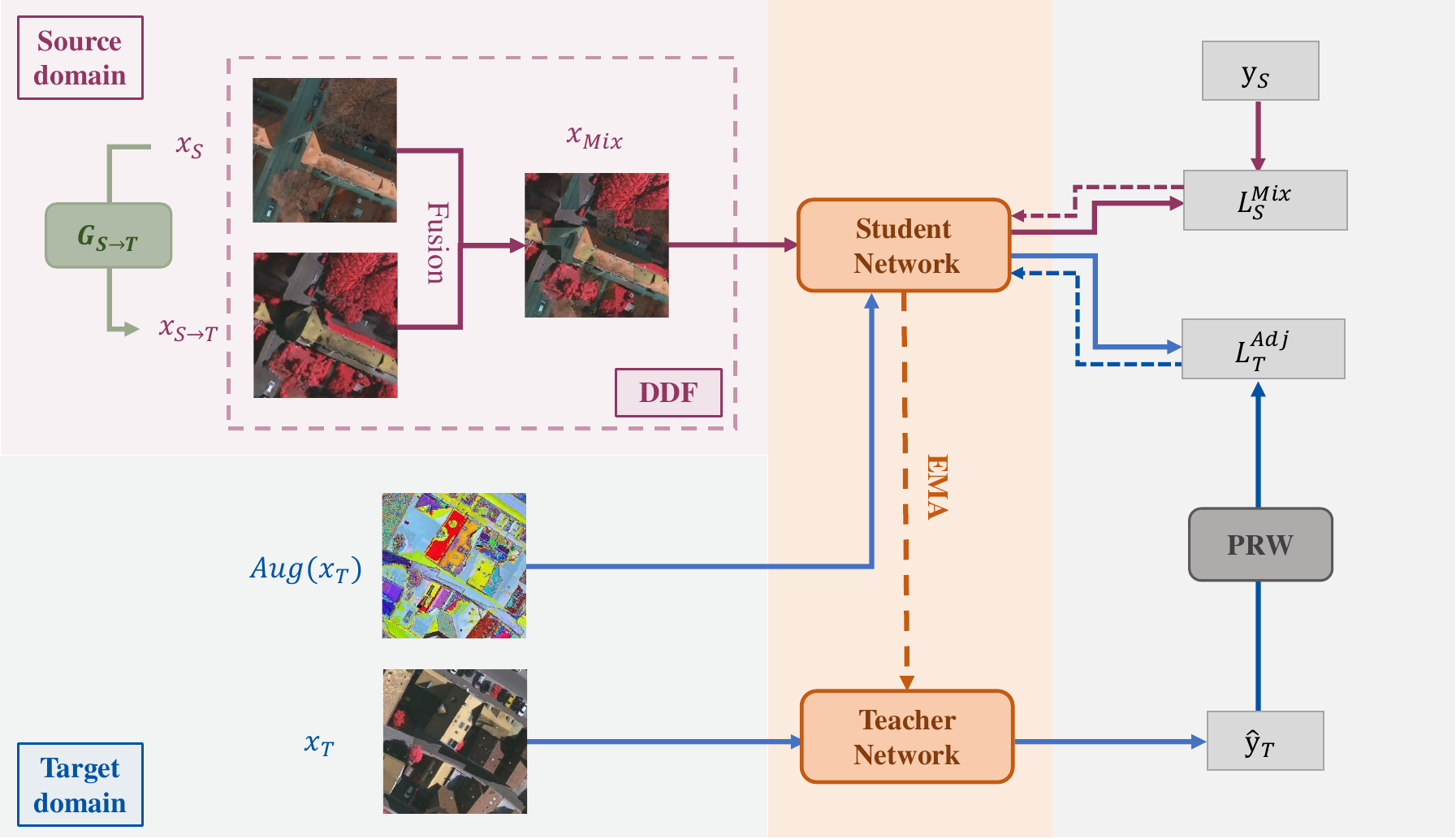}
    \centering
    \caption{The structure we propose comprises three primary elements: a self-training model that utilizes GT augmented images, a dual-domain fusion module (DDF), and a strategy (PRW) for assigning weights to pseudo-label regions. The DDF integrates the original image from the source domain with its corresponding transferred image into the student model. The teacher model assigns pseudo-labels to the target domain image, and the student model generates predictions based on these labels. We also incorporate regional adjustments to the weights of the pseudo-labels.}
    \label{fig:framework}
\end{figure*}

Self-training is a technique that involves utilizing a model trained on labeled data from a specific domain to generate pseudo-labels for data from a different domain~\cite{cbst,IAST}. These pseudo-labels are then used to retrain the model. Most methods for unsupervised domain adaptation (UDA) follow one of two approaches. The first approach involves precomputing pseudo-labels offline, training the model, and repeating this process iteratively~\cite{crst}. The second approach involves computing pseudo-labels online during training~\cite{daformer}. However, since there are inherent differences in the distribution of data between the domains, the pseudo-labels generated are likely to contain some level of noise. To mitigate the impact of incorrect labeling, pseudo-labels with a high level of confidence are often utilized.

Moreover, the application of domain mixups is also incorporated~\cite{classmix,dacs}. This approach involves merging characteristics from both the source and target domains during training phase. By doing so, it facilitates a more seamless transition between domains, reducing the model's susceptibility to domain shifts. This methodology is influenced by the concept of mixup regularization.

\section{Methodology}
This section begins by presenting a broad definition and notation of self-training for unsupervised domain adaptation, along with an explanation of its training process. Following that, we offer a comprehensive explanation of the hybrid training framework, as well as DDF and PRW modules that have been proposed.

\subsection{Self-training for UDA}

In the conventional UDA approach, a neural network is trained using a set of source domain images \(X_S = \{{x^{(i)}_S}\}^{N_S}_{i=1}\) along with their corresponding labels \(Y_S = \{{y^{(i)}_S}\}^{N_S}_{i=1}\). The objective is to fine-tune the network in order to achieve satisfactory performance on a different set of target domain images \(X_T = \{{x^{(i)}_T}\}^{N_T}_{i=1}\), where the labels \(Y_T\) are unknown.

Nevertheless, the model's accuracy drops noticeably when applied to the target domain as a result of the domain gap. Our aim is to make accurate predictions on the target domain, without having access to the target domain labels. To accomplish this, this study utilizes a self-training approach to generate pseudo-labels for the target domain. These pseudo-labels are subsequently employed to guide the training of target domain images.

The self-training method comprises a semantic segmentation network that consists of a student network \(f_\theta\) and a teacher network \(f_{\theta}^t\). The overall loss function for the student network \(f_\theta\) can be expressed as:
\begin{equation}
   L_{total}= L_S+L_T
\end{equation}

The teacher network $f_{\theta}^t$ does not update parameters with backward propagation. Generally, the weights of \(f_{\theta}^t\) are set as the exponential moving average (EMA) of the weights of \(f_\theta\) at training step t~\cite{meanteacher}. 
To avoid an increase in mis-classification probability due to erroneous results from the student model, it is recommended not to share weights with the teacher model. Instead, the EMA method can be used to aggregate information from each step, resulting in smoother output and improved pseudo-labeling quality. The teacher model parameters were updated as follows:
\begin{equation}
   \phi_t= \alpha \phi_{t-1} + (1-\alpha)\theta_t
\end{equation}

Train the student network \(f_\theta\) with a cross-entropy (CE) loss on the source domain
\begin{equation}
   L_S=-\sum y_{S}log(f_\theta(x_S))
\end{equation}

To assist the model in learning from the unlabeled target domain data, we request the teacher model \(f_{\theta}^t\) to generate pseudo-labels \(\hat{y}_T\) for the target domain data.

\begin{equation}
   \hat{y}_T=argmax(f_{\theta}^t(x_T))
\end{equation}

Assign pseudo-labels to the unlabeled data based on the model predictions and then they are used to additionally train the network \(f_\theta\) on the target domain
\begin{equation}
   L_T=-\sum \hat{y}_Tlog(f_\theta(x_{T}))
\end{equation}

\subsection{Hybrid training framework}
The main reason for incorrect pseudo-labels is the significant disparity between the two domains. The model's performance on the target domain improves as the similarity in distribution between the two domains increases. One approach, such as GT, is to transform the source domain to resemble the target domain, making the two domains appear similar. However, this method has a drawback: it does not guarantee the generation of noise-free images. If the image produces inaccurate semantic information, it can have a negative impact on the model's training. 

To tackle this issue, our training consists of two phases: image style conversion and semantic segmentation network training. In the style transfer phase, we employ a generative network consisting of a generator \(G_{S\rightarrow T}\) and a discriminator \(D_S\). \(G_{S\rightarrow T}\) translates \(X_S\) to the style of \(X_T\), resulting in \(X_{S\rightarrow T}\), which only changes the style while preserving the semantic information. \(D_S\) determines whether the image is generated, and \(G_{S\rightarrow T}\) is used to generate an image that can deceive \(D_S\). The collaboration between the two networks leads to the generation of transferred images. 

In the second stage, the semantic segmentation network includes a dual-domain fusion module, as depicted in Figure~\ref{fig:framework}. \(x_S\) and \(x_{S\rightarrow T}\) are fused through DDF to obtain the fusion image \(x_{Mix}\). The fused image retains the semantic information, so its corresponding label remains \(y_S\). \(x_{Mix}\) not only reduces the domain gap but also mitigates the impact of noise. We train the student network \(f_\theta\) using a cross-entropy loss on the source domain.


\begin{equation}
   L_S^{Mix}=-\sum y_{S}log(f_\theta(x_{Mix}))
\end{equation}

Furthermore, we will compute the quality matrix \textit{w} for pseudo-labels in order to reduce the impact of unreliable pseudo-labels. Additionally, we will give priority to detecting classes that are challenging to recognize in the image.

\begin{equation}
   L_T^{Adj}=-\sum w*\hat{y}_Tlog(f_\theta(x_{T}))
\end{equation}


Hence, the computations of losses for our framework are established as:

\begin{equation}
   L_{total}= L_S^{Mix} + L_T^{Adj}
\end{equation}

\subsection{Dual-domain image fusion}
\begin{figure}[t]
    \includegraphics[width=3.2in]{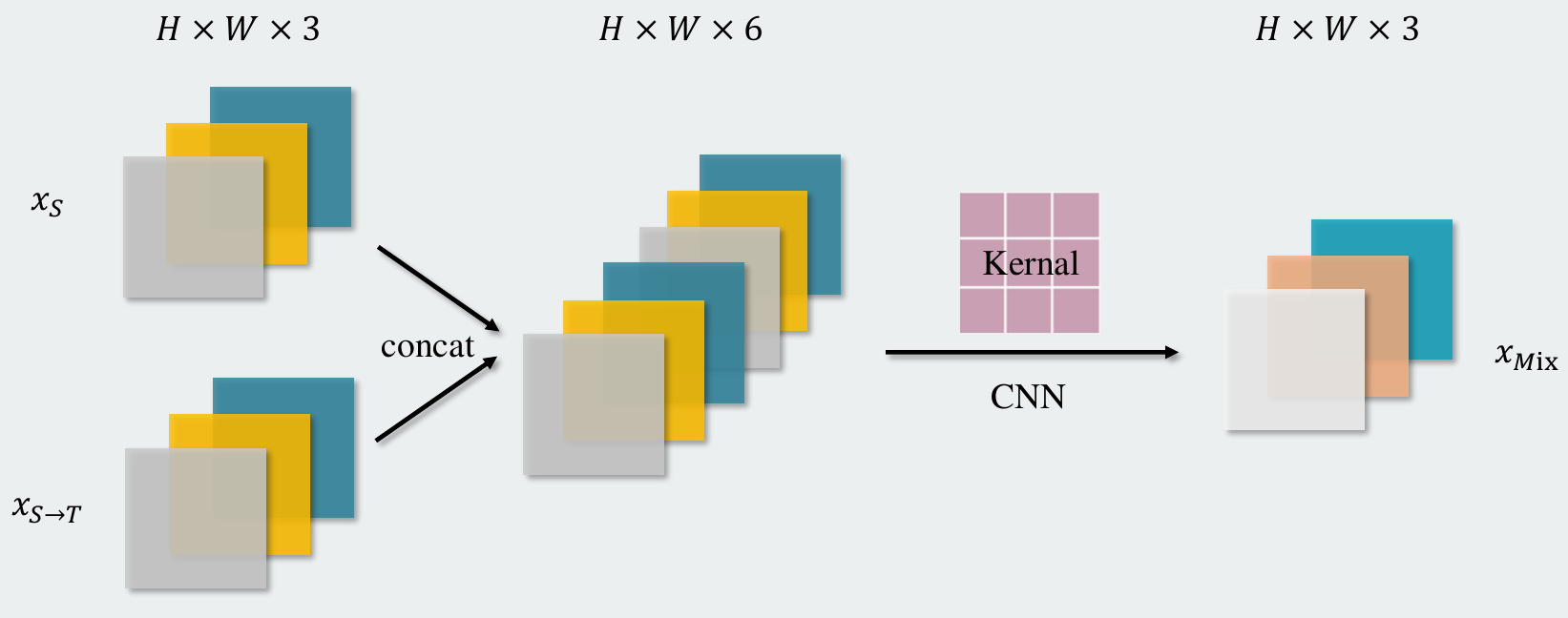}
    \centering
    \caption{Comprehensive depiction of the proposed CNN Fusion module.}
    \label{fig:cnn-fusion}
\end{figure}

\subsubsection{CNN Fusion}
To tackle the domain gap, we propose a novel image fusion module that operates in dual domains. This module effectively merges the original image with its corresponding transferred version. During the training phase, the network exclusively focuses on the source domain information when utilizing the original image. However, relying solely on the generated transferred images during training can introduce error information that may disrupt the model's training process. To overcome this challenge, we introduce a fusion image that combines the original and transferred images. This fusion image enhances the target domain information without increasing the data volume. Consequently, the network can learn both the information from both domains and the features that remain consistent across domains.

Prior to the training process, the image \(x_S\) and its corresponding transferred image \(x_{S\rightarrow T}\) should be concatenated. Both images have a size of \(H\times W\times C\).
\begin{equation}
   x_{cat}=concat(x_S,x_{S\rightarrow T}) \in \mathbb{R}^{H\times W\times 2C}
\end{equation}

Next, the input \(x_{cat}\) undergoes convolution using a convolutional network, where the convolutional kernel size is set to 3 and the number of convolutional kernels is set to 3.
\begin{equation}
  x_{Mix} = fusion\_conv(x_{cat})\in \mathbb{R}^{H\times W\times C} 
\end{equation}

The \(x_{Mix}\) obtained above, as depicted in Figure~\ref{fig:cnn-fusion}, contains both dual-domain information and preserves the semantic information of the initial image. Subsequently, this image is utilized as an enhanced source domain image for network training.

\subsubsection{Efficient Fusion}

\begin{figure}[t]
    \includegraphics[width=3.2in]{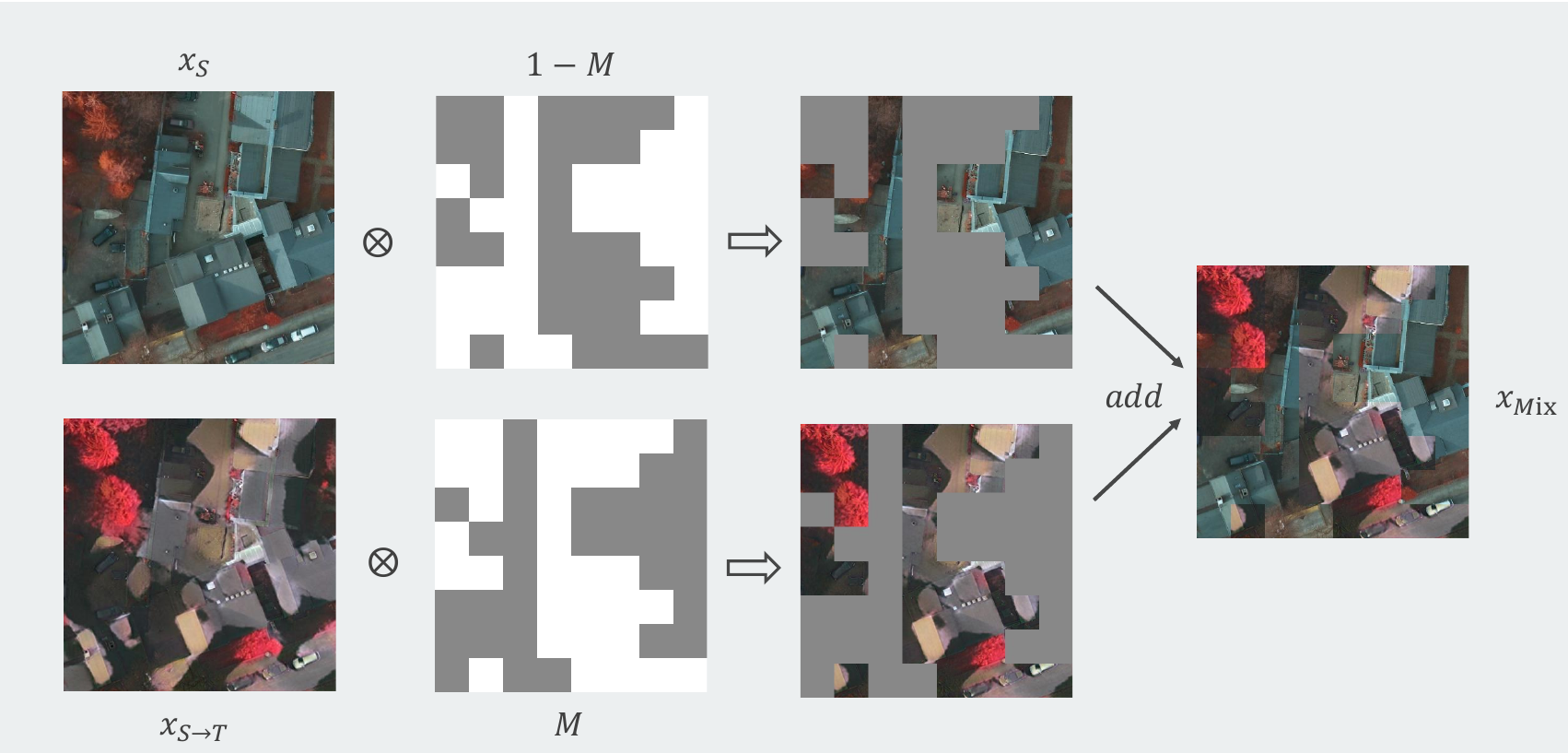}
    \centering
    \caption{Comprehensive depiction of the proposed Efficient Fusion module.}
    \label{fig:ef-fusion}
\end{figure}

Utilizing CNN fusion will result in an increase in redundant parameters and a bias towards the middle region in the fused image. In order to achieve our objective of incorporating two styles in a single image simultaneously, the most straightforward and effective approach is to merge a portion of the original image with a portion of the transferred image. Instead of randomly selecting zones, we can determine which areas to include or exclude based on the provided parameters. For this experiment, we employ two parameters: information entropy (entropy) and SND. The process is illustrated in Figure~\ref{fig:ef-fusion}.

Initially, we feed a transferred image \(x_{S\rightarrow T}\) through the student model to determine the probability of the output logits \(p_{S\rightarrow T}\). Subsequently, we divide the image into N patches of equal size \(k\times k\). We then compute the sum of the information entropy for each patch.

\begin{equation}
E[x_{S\rightarrow T}] = -\sum_{i=1}^N (p_{S\rightarrow T}^i*log(p_{S\rightarrow T}^i))
\end{equation}

Patches with lower entropy exhibit less noise in the transferred image, while patches with higher entropy tend to have more noise, which can have a detrimental effect on network training. Hence, we opt for the section of the transferred image with lower entropy and discard the higher entropy portion. By arranging the entropy values of N patches in ascending order, we can identify the lowest value t among the c\% lowest entropy blocks.
\begin{equation}
t = percentile(E[x_{S\rightarrow T}],c)
\end{equation}

Next, the patch mask M is calculated using the following algorithm:

\begin{equation}
M = \mathbbm{1}[ E[X_{S\rightarrow T}] < t],
\end{equation}
where \(\mathbbm{1}\) is an indicator function.

Choose the regions of the original image where the transferred image was not used.
In the end, we chose additional patches from both the original image and the transferred image to create a fusion image \(X_{Mix}\).
\begin{equation}
X_{Mix} = M \otimes x_{S\rightarrow T} + (1-M) \otimes x_S
\end{equation}

In addition, we can also compute the SND of the transferred image, and the procedure remains unchanged as described earlier.
\begin{equation}
SND = E[softmax(p * p^T)]
\end{equation}

\subsection{Pseudo-label regional weights}
In order to enhance the effectiveness of the pseudo-label, it is essential to evaluate its quality. This can be achieved by computing a quality matrix \textit{w} for the pseudo-labels. By setting a threshold value \(\delta\), we can determine the proportion of pixels that surpass the threshold in relation to the total number of pixels. A higher value of \textit{w} indicates a higher level of reliability for the pseudo-labels as a whole.

Nevertheless, \textit{w} treats all pseudo-labels equally. As the network undergoes multiple training iterations, it may prioritize recognizing simpler categories and overlook more complex ones. Classifying object boundaries is often challenging, so we assign higher weights to these areas. This enables us to concentrate more on the difficult categories, specifically boundaries. The objects that are easily identifiable are of lesser importance.

In order to determine the boundary area of the object, we will make use of superpixel clustering~\cite{superpixel}, which groups pixels together based on their similar features. To achieve this, we generate a binary mask \(M_b\) that represents the boundary.

\begin{equation}
M_b =  \left\{ 
    \begin{aligned}
    &0 & & x\notin boundary\quad region  \cr 
    &1 & & x\in boundary\quad region \cr 
    \end{aligned}
\right.
\end{equation}

To enhance the importance of positions within the mask region, the weights of the pseudo-labels are modified. The parameters in the original mass matrix remain unchanged for the non-boundary positions outside the mask region.

\begin{equation}
    w = w [M_b==1] + \beta , \beta \in (0,1)
\end{equation}

\section{Experiments}
\subsection{Implementation Details}

\textbf{Datasets.} Potsdam and Vaihingen are included in the ISPRS 2D open-source RS semantic segmentation benchmark dataset\footnote{https://www.isprs.org/education/benchmarks/UrbanSemLab/ semantic-labeling.aspx}. The Potsdam dataset consists of three band modes: IR-R-G, R-G-B, and IR-R-GB. The experiments use the Potsdam IR-R-G and Potsdam R-G-B, each containing 38 very high-resolution top-of-atmosphere reflectance products (VHR TOPs) with fixed dimensions of 6000 x 6000 pixels and a spatial resolution of 5 cm. The Vaihingen dataset has one band mode: IR-R-G, with 33 TOPs, each having dimensions of 2000 x 2000 pixels and a resolution of 9 cm. The images are resized to 896 x 896 pixels for Potsdam and 512 x 512 pixels for Vaihingen, resulting in a total of 1754 images for Potsdam IR-R-G and Potsdam R-G-B, and 1696 images for Vaihingen. The Vaihingen dataset is divided into training and testing sets, with 1256 images in the training set and 440 images in the testing set.

We propose two cross-domain RS semantic segmentation tasks, which are described as follows:
\begin{itemize}
    \item Potsdam IR-R-G to Vaihingen IR-R-G.
    \item    Potsdam R-G-B to Vaihingen IR-R-G.
\end{itemize}



\begin{table*}[ht]
\centering
\renewcommand\arraystretch{1.6}
\resizebox{1\linewidth}{!}{
\begin{tabular}{ccccccccccccccc}
\toprule
\multirow{2}{*}{Methods}             & \multicolumn{2}{c}{Clutter} & \multicolumn{2}{c}{Impervious surface} & \multicolumn{2}{c}{Car} & \multicolumn{2}{c}{Tree} & \multicolumn{2}{c}{Low vegetation} & \multicolumn{2}{c}{Building} & \multicolumn{2}{c}{Overall} \\ \cline{2-15} 
                                     & IoU              & F1-score            & IoU              & F1-score            & IoU       & F1-score    & IoU       & F1-score     & IoU            & F1-score          & IoU         & F1-score       & mIoU        & mF-score      \\ \hline
AdaptSegNet~\cite{adaptseg}   &      5.84& 9.01& 62.81 &76.88& 29.43& 44.83& 55.84& 71.45& 40.16 &56.87& 70.64& 82.66 &44.12 &56.95       \\ 
ProDA~\cite{proda} &  3.99& 8.21 &62.51 &76.85& 39.20& 56.52& 56.26& 72.09 &34.49 &51.65 &71.61& 82.95 &44.68& 58.05    \\ 
Bai's~\cite{Bai} &       19.60& 32.80& 65.00& 78.80 &39.60 &56.70& 54.80& 70.80& 36.20& 53.20 &76.00& 86.40 &48.50 &63.10 \\
Zhang's~\cite{zhang} &   20.71& 31.34& 67.74& 80.13& 44.90& 61.94& 55.03 &71.90 &47.02 &64.16 &76.75 &86.65& 52.03& 66.02  \\
Wang's~\cite{Wang2023} &   21.85&35.87&76.58&86.73&35.44&52.33&55.22&71.15&49.97&66.64&82.74&90.56&53.63&67.21\\
CIA-UDA~\cite{CIA} &  27.80&  43.51&  63.28&  77.51&  52.91 & 69.21 & 64.11&  78.13&  48.03&  64.90&  75.13&  85.80&  55.21&  69.84  \\
ResiDualGAN~\cite{residualgan} &11.64 &18.42&72.29 &83.89& 57.01 &72.51 &63.81& 77.88& 49.69& 66.29 &80.57 &89.23& 55.83 &68.04 \\  \hline
ST-DASegNet~\cite{st-da} &   \textbf{67.03}& \textbf{80.28} &74.43& 85.36 &43.38& 60.49& 67.36 &80.49 &48.57 &65.37 &85.23 &92.03& 64.33& 77.34\\
DAFormer*~\cite{daformer} &   41.21& 58.37& \underline{77.95}& \underline{87.61}& \underline{62.21}& \underline{76.70} &\textbf{70.80}& \textbf{82.90}& \underline{52.38} &\underline{68.75} &\textbf{87.44}&\textbf{ 93.30} &\underline{65.33}& \underline{77.94} \\
 \hline
Ours & \underline{59.69}&\underline{74.76}&\textbf{79.72}&\textbf{88.72}&\textbf{63.04}&\textbf{77.33}&\underline{70.03}&\underline{82.37}&\textbf{52.94}&\textbf{69.23}&	\underline{86.70}&\underline{92.88}& \textbf{68.69} &\textbf{80.88} \\
\bottomrule
\multicolumn{4}{l}{\small ``*" are our re-implemented version for RS images.}
\end{tabular}}
\caption{The quantitative results of the cross-domain semantic segmentation from Potsdam IR-R-G to Vaihingen IR-R-G.}
\label{tab:p2v}
\end{table*}

\begin{table*}[hbtp]
\centering
\renewcommand\arraystretch{1.6}
\resizebox{1\linewidth}{!}{
\begin{tabular}{ccccccccccccccc}
\toprule
\multirow{2}{*}{Methods}             & \multicolumn{2}{c}{Clutter} & \multicolumn{2}{c}{Impervious surface} & \multicolumn{2}{c}{Car} & \multicolumn{2}{c}{Tree} & \multicolumn{2}{c}{Low vegetation} & \multicolumn{2}{c}{Building} & \multicolumn{2}{c}{Overall} \\ \cline{2-15} 
                                     & IoU              & F1-score            & IoU              & F1-score            & IoU       & F1-score    & IoU       & F1-score     & IoU            & F1-score          & IoU         & F1-score       & mIoU        & mF-score      \\ \hline
AdaptSegNet~\cite{adaptseg}        &  6.49 &9.82& 55.70 &71.24& 33.85 &50.05& 47.72& 64.31 &22.86& 36.75 &65.70& 79.15& 38.72 &51.89    \\ 
ProDA~\cite{proda} &    2.39& 5.09& 49.04& 66.11& 31.56& 48.16 &49.11 &65.86& 32.44 &49.06& 68.94 &81.89 &38.91& 52.70     \\ 
Bai's~\cite{Bai} &     10.80&  19.40 & 62.40&  76.90&  38.90 & 56.00&  53.90 & 70.00 & 35.10&  51.90&  74.80&  85.60&  46.00&  60.00\\
ResiDualGAN~\cite{residualgan} &9.76 &16.08 &55.54& 71.36& 48.49 &65.19& 57.79 &73.21& 29.15& 44.97& 78.97 &88.23& 46.62& 59.84 \\
Zhang's~\cite{zhang}&   12.38&  21.55&  64.47&  77.76&  43.43&  60.05 & 52.83&  69.62&  38.37 & 55.94 & 76.87 & 86.95 & 48.06&  61.98 \\
Wang's~\cite{Wang2023} & 12.61&22.39& 73.80 &\textbf{84.92}& 43.24& 60.38 &44.41 &61.50 &43.27& 60.40& 83.76& 91.16& 50.18 &63.46\\
CIA-UDA~\cite{CIA} &    13.50& 23.78 &62.63 &77.02& 52.28& 68.66& 63.43& 77.62 &33.31& 49.97 &79.71 &88.71& 50.81& 64.29 \\  \hline
ST-DASegNet~\cite{st-da} &     36.03 &50.64&68.36& 81.28 &43.15 &60.28& 64.65& 78.31 &34.69& 47.08& 84.09 &91.33 &55.16& 68.15 \\
DAFormer*~\cite{daformer} &   \underline{39.66} & \underline{56.79} & \underline{69.98} & 82.34 &\underline{58.01} & \underline{73.43} &\underline{69.21}& \underline{81.81}& \underline{45.76 }& \underline{62.79} &\textbf{85.82} &\textbf{92.37}& \underline{61.41} &\underline{74.92} \\
 \hline
Ours &  \textbf{ 54.38}   &  \textbf{75.75 } &\textbf{74.79}&    \underline{84.46}  &  \textbf{ 60.32}  &  \textbf{75.17}  &   \textbf{ 70.19} &  \textbf{83.86 }  &   \textbf{50.01 }  &  \textbf{69.20 }    &   \underline{84.59}  &  \underline{92.17}  & \textbf{65.71} & \textbf{80.10}\\
\bottomrule
\multicolumn{4}{l}{\small ``*" are our re-implemented version for RS images.}
\end{tabular}}
\caption{The quantitative results of the cross-domain semantic segmentation from Potsdam R-G-B to Vaihingen IR-R-G.}
\label{tab:prgbv}
\end{table*}

\textbf{Network Architecture and training.} We adopt SegFormer~\cite{segformer} as the architecture, which is pretrained on ImageNet-1k. To train the network, we employ AdamW~\cite{adamw} optimizer with a learning rate of \(6\times 10^{-5}\) for the encoder and \(6\times 10^{-4}\) for the decoder. A weight decay of \(0.01\) is applied, along with a linear learning rate warmup for 1.5k steps, followed by linear decay. The data augmentation used is consistent with DACS~\cite{dacs}. All models are trained on an NVIDIA Tesla V100.

\textbf{Evaluation Metric.} To enable easy comparison with other methods, this research employs the widely used evaluation metrics of mIoU and F1-score for semantic segmentation. The IoU is calculated for each category using the formula \(IoU = A\cap B / A \cup B\). The mIoU represents the average IoU score across all classes. The F1-score is defined as \(F_{1-score} = (2\cdot Precision\cdot Recall) / (Precision + Recall)\).

\subsection{Comparison with SOTA methods}
\subsubsection{Comparison experiments from Potsdam IR-R-G to Vaihingen IR-R-G}
In this experiment, we use Potsdam IR-R-G images as the source domain and Vaihingen IR-R-G images as the target domain. The model is trained on 1764 annotated images from Potsdam and 1296 unannotated training images from Vaihingen. The evaluation is conducted on 440 test images from Vaihingen. The comparison of results is shown in Table~\ref{tab:p2v}. The first method is based on DeepLabV3, while the second method is based on SegFormer. Our method outperforms the SegFormer-based method (DAFormer) with a 3.36\% improvement in mIoU and a 2.94\% improvement in mF-score. Specifically, we achieve better results in the categories of ``Impervious surface", ``Car", and ``Low vegetation". The difference between our results and the best results in the ``Tree" and ``Building" categories is also minimal.

\begin{table*}[t]
\centering
\renewcommand\arraystretch{1.6}
\resizebox{1\linewidth}{!}{
\begin{tabular}{ccccccccccccccc}
\toprule
\multirow{2}{*}{Methods}             & \multicolumn{2}{c}{Clutter} & \multicolumn{2}{c}{Impervious surface} & \multicolumn{2}{c}{Car} & \multicolumn{2}{c}{Tree} & \multicolumn{2}{c}{Low vegetation} & \multicolumn{2}{c}{Building} & \multicolumn{2}{c}{Overall} \\ \cline{2-15} 
                                     & IoU              & F1-score            & IoU              & F1-score            & IoU       & F1-score    & IoU       & F1-score     & IoU            & F1-score          & IoU         & F1-score       & mIoU        & mF-score      \\ \hline
Base   &  30.22&46.42&77.42&87.27&60.47&75.37&70.63&82.79&51.76&68.21&87.06&93.08&62.93&75.52   \\ 
Base +DDF (CNN)              &  55.95      & 71.75       & 77.53     & 87.35    &   56.45   & 72.16     &  71.01    &   83.05    &  52.60       &     68.93    &  85.57    &   92.22    &   66.52  &   79.24   \\ 
Base +DDF (Efficient)       & 54.00           & 70.13              & 79.16          & 88.37              & 63.14    & 77.40      & 70.64    & 82.80       & 53.89        & 70.04      & 86.53       & 92.78          & 67.89      & 80.25        \\ 
Base +PRW & 52.62            & 68.96               & 79.27            & 88.44               & 62.29     & 76.77       & 71.39     & 83.31        & 55.29          & 71.21             & 86.23       & 92.61          & 67.85       & 80.21         \\ 
Base +DDF +PRW       & 59.69&74.76&79.72&88.72&63.04&77.33&70.03&82.37&52.94&69.23&	86.70&92.88&68.69&80.88               \\ \bottomrule
\end{tabular}}
\caption{Ablation experiments of the cross-domain semantic segmentation from Potsdam IR-R-G to Vaihingen IR-R-G}
\label{tab:pv-ablation}
\end{table*}

\begin{table*}
\centering
\renewcommand\arraystretch{1.6}
\resizebox{1\linewidth}{!}{
\begin{tabular}{ccccccccccccccc}
\toprule
\multirow{2}{*}{Methods}             & \multicolumn{2}{c}{Clutter} & \multicolumn{2}{c}{Impervious surface} & \multicolumn{2}{c}{Car} & \multicolumn{2}{c}{Tree} & \multicolumn{2}{c}{Low vegetation} & \multicolumn{2}{c}{Building} & \multicolumn{2}{c}{Overall} \\ \cline{2-15} 
                                     & IoU              & F1-score            & IoU              & F1-score            & IoU       & F1-score    & IoU       & F1-score     & IoU            & F1-score          & IoU         & F1-score       & mIoU        & mF-score      \\ \hline
Base                  & 14.37       & 25.14         & 69.20            & 81.79               & 61.90     & 76.47       & 69.51     & 82.02        & 40.17          & 57.32             & 85.95       & 91.86          & 56.68       & 69.10          \\
Base +DDF (CNN)             & 46.34       & 63.34         & 75.25            & 85.88               & 58.93     & 74.16       & 72.17     & 83.83        & 55.29          & 71.21             & 85.60       & 92.24          & 65.60       & 78.44         \\
Base +DDF (Efficient)         & 47.89       & 64.76         & 75.45            & 86.01               & 59.94     & 74.95       & 69.48     & 81.99        & 47.93          & 64.80              & 85.30        & 92.07          & 64.33       & 77.43         \\
Base +PRW & 40.91       & 58.06         & 76.22            & 86.51               & 60.68     & 75.53       & 69.29     & 81.86        & 51.14          & 67.68             & 85.44      & 92.15       & 63.95       & 76.96         \\
Base +DDF +PRW        &   54.38   &  75.75  & 74.79 &    84.46  &   60.32  &  75.17  &    70.19 &  83.86   &   50.01   &  69.20  &   84.59  &  92.17  & 65.71 & 80.10    \\ 
\bottomrule
\end{tabular}}
\caption{Ablation experiments of the cross-domain semantic segmentation from Potsdam R-G-B to Vaihingen IR-R-G}
\label{tab:prgbv-ablation}
\end{table*}

\subsubsection{Comparison experiments from Potsdam R-G-B to Vaihingen IR-R-G}
In this experiment, we use Postsdam R-G-B images as the source domain and Vaihingen IR-R-G images as the target domain. The model is trained using 1764 annotated images from Potsdam and 1296 unannotated training images from Vaihingen. The evaluation is conducted on 440 test images from Vaihingen. This task is more challenging than the ``Potsdam IR-R-G to Vaihingen IR-R-G" task due to the differences in sensors, which results in a larger gap between the two domains. However, our results show significant improvements in this task. The comparative results can be seen in Table~\ref{tab:prgbv}. Notably, our approach achieves state-of-the-art (SOTA) results and performs the best in five categories. It also demonstrates exceptional performance in the ``Clutter" category, surpassing the second-ranked method by a significant margin. Specifically, our method achieves a 14.72\% improvement in IoU value and an 18.98\% enhancement in F1-score.

\subsection{Visualization}
\begin{figure}[t]
    \includegraphics[width=3in]{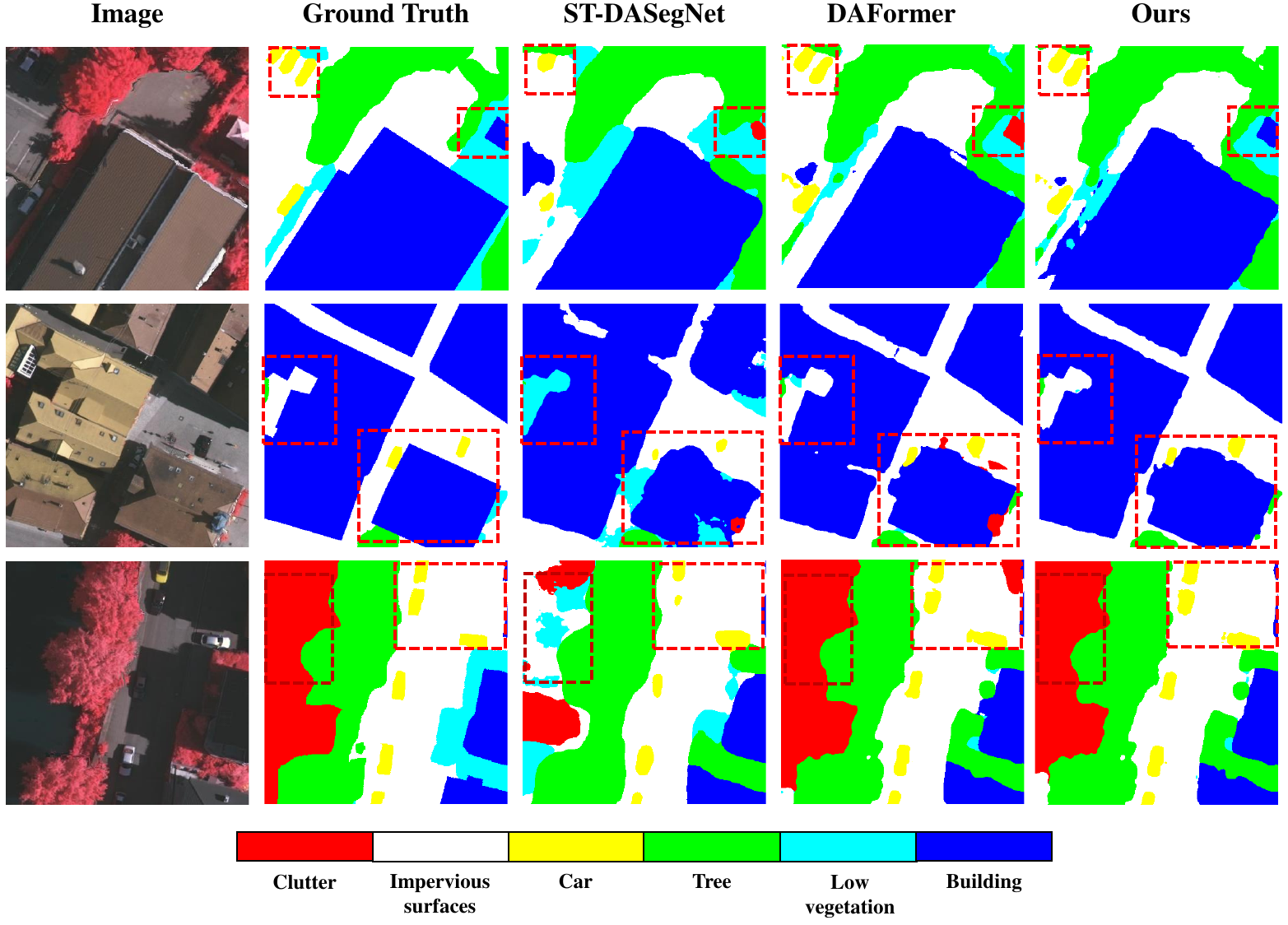}
    \centering
    \caption{The qualitative visualization of the cross-domain semantic segmentation from Potsdam IR-R-G to Vaihingen IR-R-G.}
    \label{fig:Visualization}
\end{figure}

This subsection primarily focus on the visualization results in our experimental analysis. Our method surpasses existing methods in various aspects. The results depicted in Figure~\ref{fig:Visualization} demonstrate that our model accurately identifies and segments complex mixed objects, thereby reducing the instances where other classes are mistakenly identified as the ``Clutter" category.
Additionally, our method significantly enhances the performance in the ``Impervious surface" category. Compared to other methods for comparison, our approach better captures the surface's details and textures, resulting in sharper segmentation boundaries with minimal blurring or misalignment. In terms of the ``Car" category segmentation, our method performs well by successfully distinguishing cars from their surroundings. 

By testing our method on images from different scenes, we not only achieve more accurate results but also obtain clearer boundaries. The sensitivity of our method to edge detection reduces the likelihood of misclassification.

The visualization outcomes confirm the outstanding performance of our proposed technique in the three main categories: ``Clutter", ``Impervious surface", and ``Car". These outcomes showcase the efficiency of our approach and establish a practical basis for implementing semantic segmentation in remote sensing images. These benefits will contribute to enhanced precision and dependability in analyzing remote sensing images for future practical applications.

\subsection{Ablation Study}
In the ablation experiment, we conducted a comparison of multiple variants in order to assess the impact of each component on performance. The results can be seen in Table~\ref{tab:pv-ablation} and Table~\ref{tab:prgbv-ablation}. The DDF module, which stands for dual-domain image fusion, and the PRW approach, which stands for pseudo-label regional weight, were examined. The network structure remained consistent throughout the training process for all tasks. The ``Base" task was identical to the other tasks, except for the absence of the aforementioned modules. The DDF module displayed significantly better performance in the ``clutter" class, whether integrated with CNN Fusion or Efficient Fusion. This can be attributed to the fact that the DDF module effectively reduces the domain gap by fusing information from both domains, leading to improved performance in the most challenging class, ``Clutter". On the other hand, the PRW approach primarily focuses on challenging classes. As a result, the addition of PRW greatly improved the performance of the ``Clutter" and ``Low vegetation" classes, which had low performance in the ``Base" task. Furthermore, due to the use of different sensors in the Potsdam R-G-B and Vaihingen IR-R-G datasets, the difference between the source and target domains was more pronounced, making the task of ``Potsdam R-G-B to Vaihingen IR-R-G" more difficult. However, our proposed module demonstrated significant advancements in this task, effectively minimizing the domain gap, particularly in domains with substantial differences.

\section{CONCLUSION}
In this paper, we present a novel approach for domain-adaptive semantic segmentation of remote sensing images. Our method consists of three main elements: a hybrid training approach, dual-domain image fusion, and regional weight pseudo-labeling. The hybrid training strategy enhances the performance of self-training by using augmented GT images. The dual-domain image fusion strategy generates intermediate-domain information and reduces the discrepancy between different domains. The regional weighting of pseudo-labels assigns higher weights to categories that are more difficult to identify, leading to significant improvements in segmentation accuracy for those categories. 
To demonstrate the effectiveness of the proposed approach, extensive benchmark experiments and ablation studies are conducted on the ISPRS Vaihingen and Potsdam datasets. Moving forward, our future research endeavors will be directed towards further investigating the use of intermediate domain information to minimize domain gap.

\clearpage






\bibliographystyle{named}
\bibliography{ijcai24}

\end{document}